\documentclass{article}

\usepackage[nonatbib,preprint]{neurips_2025}

\usepackage[utf8]{inputenc} 
\usepackage[T1]{fontenc}    
\usepackage[hidelinks]{hyperref}       
\usepackage{url}            
\usepackage{booktabs}       
\usepackage{amsfonts}       
\usepackage{nicefrac}       
\usepackage{microtype}      
\usepackage[table]{xcolor}         
\usepackage{graphicx}
\usepackage{amsmath}
\usepackage{array} 
\usepackage[table]{xcolor} 
\usepackage{colortbl}      
\usepackage{siunitx}
\newcolumntype{H}{>{\setbox0=\hbox\bgroup}c<{\egroup}@{}}

\title{Flash-VL 2B: Optimizing Vision-Language Model Performance for Ultra-Low Latency and High Throughput}

\author{%
  Bo Zhang 
  \And
  Shuo Li
  \And 
  Runhe Tian
  \And
  Yang Yang
  \And
  Jixin Tang
  \And
  Jinhao Zhou
  \And
  Lin Ma \\
  \AND {} \\
  Meituan \\
}

\begin{document}

\maketitle

\begin{abstract}
In this paper, we introduce Flash-VL 2B, a novel approach to optimizing Vision-Language Models (VLMs) for real-time applications, targeting ultra-low latency and high throughput without sacrificing accuracy. Leveraging advanced architectural enhancements and efficient computational strategies, Flash-VL 2B is designed to maximize throughput by reducing processing time while maintaining competitive performance across multiple vision-language benchmarks. Our approach includes tailored architectural choices, token compression mechanisms, data curation, training schemes, and a novel image processing technique called implicit semantic stitching that effectively balances computational load and model performance. Through extensive evaluations on 11 standard VLM benchmarks, we demonstrate that Flash-VL 2B achieves state-of-the-art results in both speed and accuracy, making it a promising solution for deployment in resource-constrained environments and large-scale real-time applications.
\end{abstract}

\section{Introduction}

The rapid advancements in Vision-Language Models (VLMs) have led to significant breakthroughs in multimodal understanding, enabling systems to integrate visual and textual information for a variety of applications such as image captioning, visual question answering, and cross-modal retrieval. Recent models, including LLaVA \cite{liu2023llava,liu2023improvedllava}, Qwen-VL \cite{Qwen-VL,Qwen2VL}, Intern-VL \cite{chen2024internvl,chen2024far}, and MobileVLM \cite{chu2023mobilevlm,chu2024mobilevlm}, have pushed the boundaries of performance by scaling model sizes and refining architectures. These models have set new benchmarks, demonstrating significant accuracy improvements across multiple vision-language tasks. However, as model complexity increases, there is a critical need to address the trade-off between performance and computational efficiency, particularly for real-time and resource-constrained environments.

Despite their remarkable accuracy, existing state-of-the-art VLMs often face challenges in deployment due to their high latency and computational demands, limiting their applicability in applications requiring fast and scalable inference. For example, recent VLMs support dynamic resolution either by splitting images into finer patches~\cite{chen2024internvl} or resizing to the nearest size to facilitate downsampling~\cite{Qwen2VL}, hence substantially increasing the number of visual tokens for LLMs to process. In practice, both approaches increase visual encoding latency as well as the duration of the decoding stage. DeepSeek-VL2~\cite{wu2024deepseekvl2} leverages mixture-of-experts as LLMs to enhance performance, however the sparse MoE architecture naturally requires more memory access, which hinders the inference speed.


\begin{figure}[htbp]
    \centering
    \begin{minipage}[t]{0.48\textwidth}
    \centering
    \includegraphics[width=6cm]{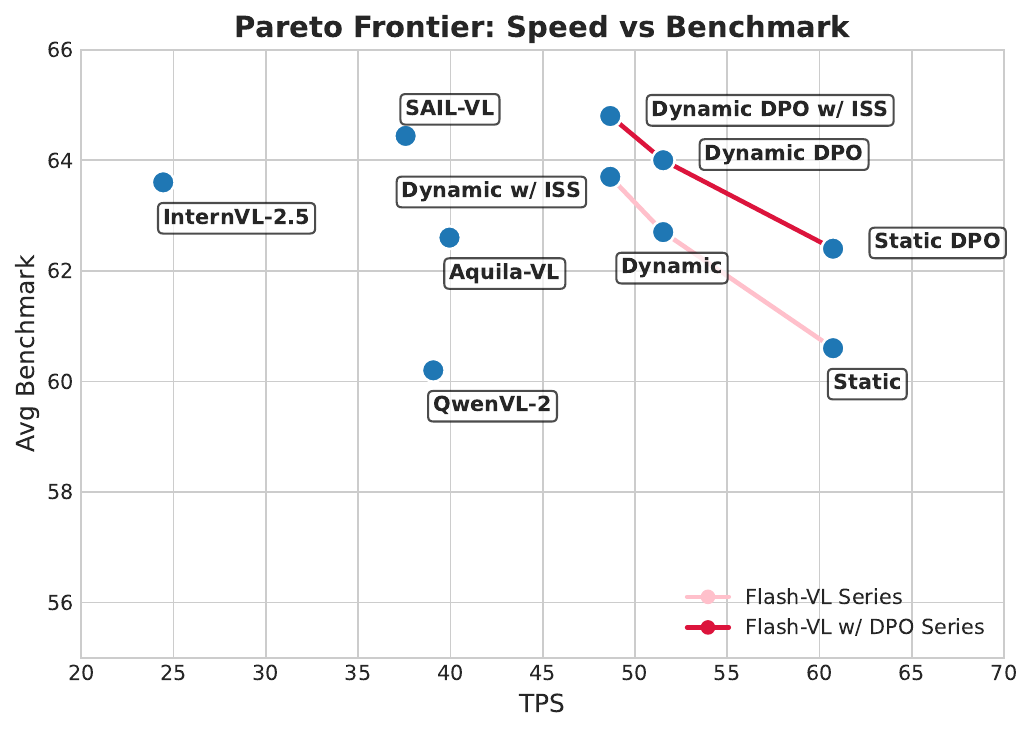}
    \end{minipage}
    \begin{minipage}[t]{0.48\textwidth}
    \centering
    \includegraphics[width=7cm]{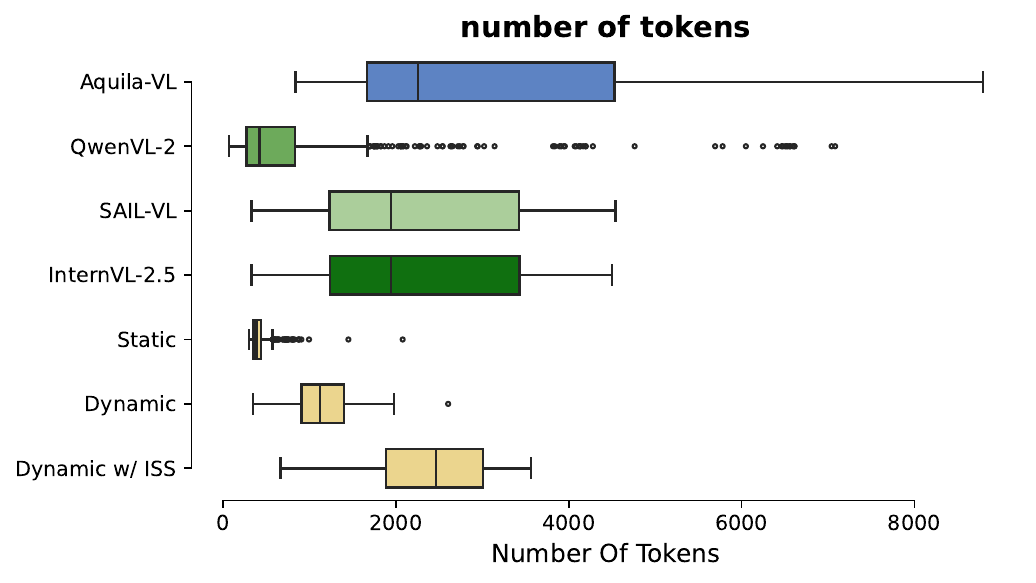}
    \end{minipage}
    \caption{\textbf{Left:} Pareto front of VLMs in terms of the average accuracy on 11 standard VLM benchmarks and the TPS (tokens per second). \textbf{Right:} The distribution of the input token numbers processed by the LLM component of each VLM given various image sizes of the MMMU dataset.}
    \label{fig:pareto}
\end{figure}

To bridge this gap, we propose Flash-VL 2B, a new VLM architecture designed to achieve ultra-low latency and higher throughput without sacrificing accuracy. By incorporating cutting-edge architectures, novel tiling strategies, token compression techniques, and advanced data and training schemes, Flash-VL 2B aims to provide a more efficient alternative to current VLMs, particularly for applications that require fast real-time performance and scalability. In this work, we demonstrate that Flash-VL 2B not only delivers state-of-the-art accuracy on common VLM benchmarks but also provides significant improvements in speed and throughput (up to 60.73 tokens/s in Figure~\ref{fig:pareto} left), making it a promising candidate for deployment in both cloud and edge environments. Our Flash-VL-2B series has a higher compression on the number of total tokens, faster than its counterparts, while being better or on par in average performance on 11 standard VLM benchmarks.

\section{Related Work}
The prominent VLMs so far generally adopt a ViT-Adapter-LLM architecture disseminated by LLaVA~\cite{liu2023llava}. For an efficient design of VLMs, the choice of visual encoders~\cite{dosovitskiy2020image} and LLMs incurs a critical trade-off between strength and efficiency.

\textbf{Vision Encoders.} Contrastive language-image pretraining~\cite{radford2021learning} has become the most powerful method for visual representation learning. By training over 400M image-text pairs with weak language supervision, CLIP~\cite{radford2021learning} unlocks competitive ImageNet zero-shot accuracy and many other downstream tasks. Open-CLIP~\cite{gadre2023datacomp} lays more importance on data curation, releasing a massive filtered high-quality multi-modal dataset called DataComp-1B.  Eva-CLIP~\cite{sun2023eva} further boosts the performance by scaling training data to have 9B samples. Intern-ViT~\cite{chen2024internvl} scales ViTs to 6B to parallel the number of parameters of LLMs. DFN~\cite{fang2023data} emphasizes a filtering strategy of data curation and releases a new dataset DFN-5B filtered out of 12.8B CommonPool. Most recently, SigLIP~\cite{zhai2023sigmoid} enhances CLIP by transforming the softmax-based contrastive loss to a sigmoid one. Its successor SigLIP2~\cite{tschannen2025siglip} extends the training objectives to incorporate caption-based pretraining~\cite{wan2024locca} and self-distillation as in DINOv2~\cite{oquab2023dinov2}. AIM~\cite{el2024scalable,fini2024multimodal} series is solely built on an autoregressive objective where the model generates the sequence of raw image patches and text tokens (added in AIMv2 by pairing a multimodal decoder). Among all, the SigLIP series achieves a nice trade-off between inference cost and performance, offering more compact but competitive models.

\textbf{Lightweight LLMs.} LLaMA series~\cite{touvron2023llama,touvron2023llama2,grattafiori2024llama} provides a wide range of language models, where the smallest one of earlier versions has 7B parameters. To achieve ultra-low latency, pruning methods like LLM-Pruner~\cite{ma2023llmpruner} and ~\cite{muralidharan2024compact} are introduced to nearly compress LLMs by half of the parameters. In contrast, others train smaller models from scratch, which has been proven to have higher compactness in terms of accuracy at the same scale. Phi series~\cite{gunasekar2023textbooks,li2023textbooks,abdin2024phi} targets phone use cases by training light-weight models on the high-quality textbook corpus. Gemma 2~\cite{team2024gemma2} utilizes distillation to obtain smaller models. MiniCPM~\cite{hu2024minicpm} studies the scaling law in extremely small LLMs and proposes a warmup-stable-decay training strategy for better convergence. Qwen~\cite{bai2023qwen,yang2024qwen2} ships a rich configuration of models equipped with architectural upgrades and trained on larger-scale data (up to 18T training tokens), whose smallest version serves as a very competitive base model for building up light-weight VLMs.

\textbf{Efficient VLMs.} LLaVA series~\cite{liu2023llava,liu2023improvedllava} illustrates a popular paradigm and also a training recipe for efficient VLMs. MobileVLM~\cite{chu2023mobilevlm,chu2024mobilevlm} aims to increase the inference speed by compressing visual tokens via a light-weight adapter. Phi-4~\cite{abouelenin2025phi} and Gemma 3~\cite{team2025gemma} are the multi-modal follow-ups built on their previous lightweight LLMs. NVILA~\cite{liu2024nvila} undertakes a ``scale-then-compress" approach, which allows efficient processing of high-resolution images and long videos.  FastVLM~\cite{vasu2024fastvlm} makes use of a hybrid vision encoder to significantly reduce encoding time for high-resolution
images. LLaVA-OneVision~\cite{li2024llava} comes with efficient VLMs that handle images, texts, and videos. Qwen-VL2~\cite{Qwen2VL} proposes a dynamic resolution mechanism to process images of arbitrary size, it also unifies image and video representation by integrating 3D convolutions and multi-modal RoPE~\cite{su2024roformer}. Aquila-VL~\cite{gu2024infinitymmsc} curates a 40M open-sourced multimodal dataset that significantly increases the training efficiency. Lately, SAIL-VL~\cite{dong2025scalable} enhances the data pipeline to produce scalable data where models can be trained based on the sample complexity via curriculum learning.


\section{Flash-VL}

\subsection{The VLM Architecture}

\begin{figure}
    \centering
    \includegraphics[width=\textwidth]{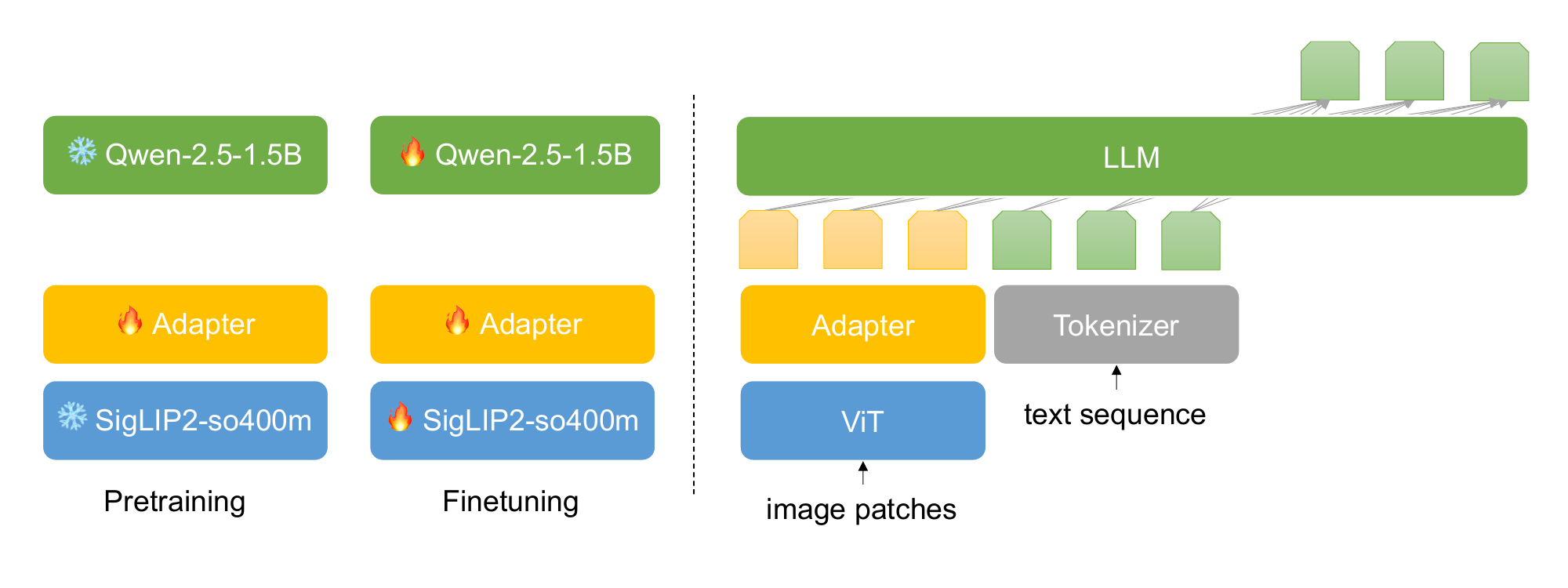}
    \caption{The Flash-VL architecture adopts a ViT-Adapter-LLM paradigm trained in multi-stages, striking an outstanding trade-off between accuracy and speed.}
    \label{fig:architecture}
\end{figure}

Following LLaVA~\cite{liu2023llava}, we choose a conventional Vit-Adapter-LLM setup for constructing our VLM. With inference requirements and compute budget in mind, we select SigLIP2-so400m-patch16-512~\cite{tschannen2025siglip} by default for the vision encoder and Qwen-2.5-1.5B-Instruct~\cite{qwen2.5} as our language model. These two major components are connected by a lightweight adapter with token compression done by pixel shuffling, as depicted in Figure~\ref{fig:architecture}.

\textbf{Visual Encoder.} For ultra-low latency and high throughput, we refrain from the common dynamic tiling strategy ~\cite{chen2024far,hu2024minicpm} as it substantially induces computational costs. Among all available visual encoders, SigLIP2~\cite{tschannen2025siglip} features fewer parameters at a lower fixed image resolution while trained at dynamic resolution scales. We compare SigLIP2 and AIMv2~\cite{fini2024multimodal} later in Table~\ref{tab:vit-comp}, manifesting the superiority of SigLIP2 in visual representation capacity. For each input image, we resize to 512$\times$512 and slice out 32$\times$32 patches each of size 16$\times$16, yielding 1024 visual tokens in total. 

\textbf{Adapter.} To compress the number of visual tokens for efficiency, we adopt pixel shuffling~\cite{chen2024far} to swap every other two pixels in the $H$ and $W$ dimensions to the hidden dimension, effectively reducing the number of tokens by four times to have 256 tokens. Beyond that, we append layer normalization and linear layers with intermediate activations to align with the LLM. Formally, for given visual feature $x$, we process it as follows,

\begin{align}
    x &= PixelShuffle(x) \\
    x &= LayerNorm(x) \\
    x &= GELU(Linear(x)) \\
    x &= GELU(Linear(x)) \\
    x &= Linear(x) 
\end{align}
The comparison of this adapter with the conventional two MLP layers is in Table~\ref{tab:adp-comp} of Section~\ref{app:ablation-adp}.

\textbf{Language Model.} Among all dense lightweight models below 2B parameters, Qwen-2.5-1.5B-Instruct~\cite{qwen2.5} boasts the best overall performance on various language tasks. It ships a potent causal LLM out-of-box with advanced architectural upgrades like SwiGLU~\cite{shazeer2020glu}, RoPE~\cite{su2024roformer}, RMSNorm~\cite{zhang2019root}, and tied word embeddings~\cite{press2016using}. It also enables a large vocabulary of 151936 and a context length of 32K (including an 8K output length). We compare it with other LLM candidates later in Table~\ref{tab:llm-comp} of Section~\ref{app:ablation-llm}.

\begin{figure}[ht]
    \centering
    \includegraphics[width=1.0\textwidth]{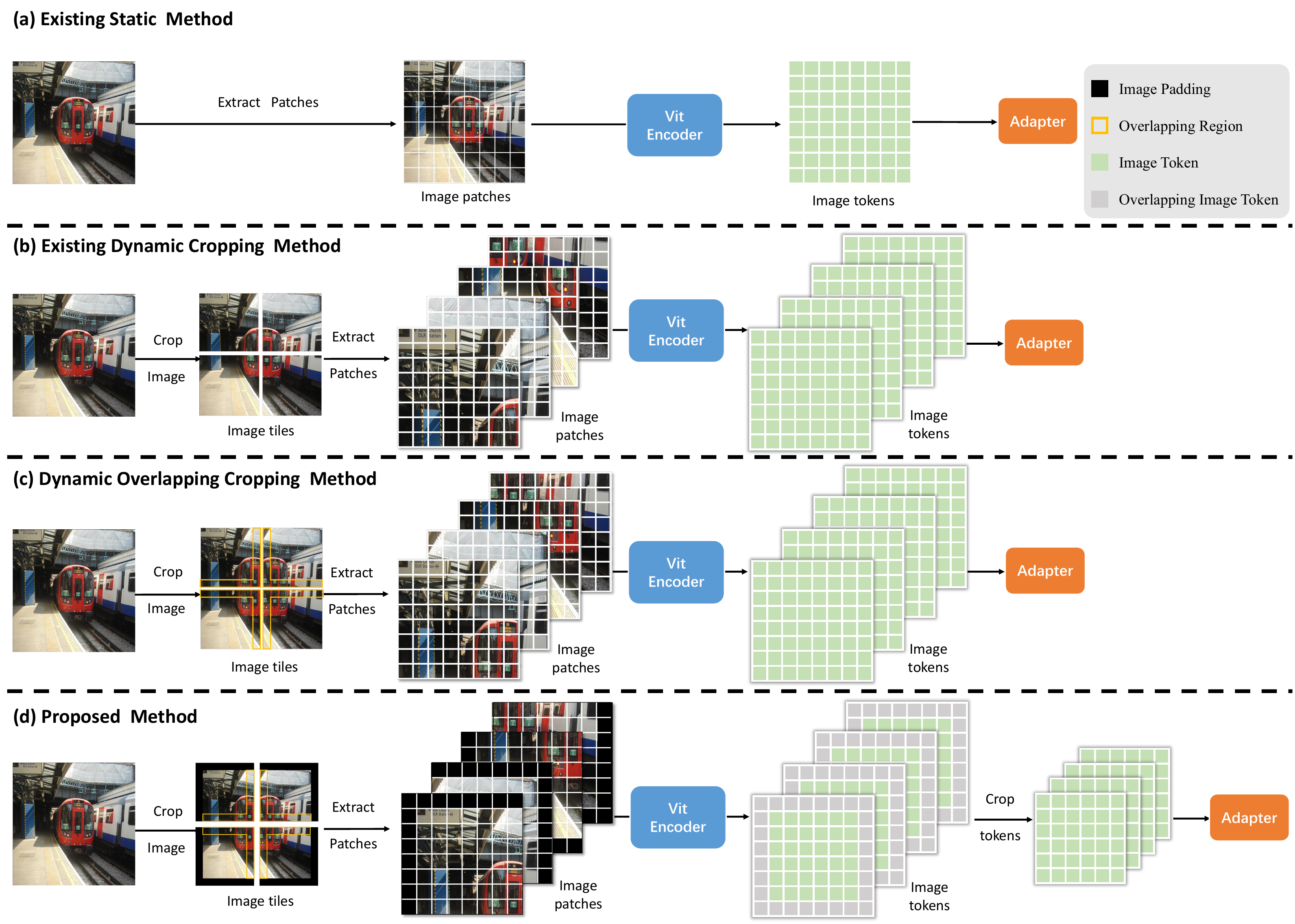}
    \caption{Image processing strategies in VLMs. (a) Static Method (b) Dynamic Cropping Method (c) Dynamic Overlapping Cropping Method (d) The Proposed Implicit Semantic Stitching.}
    \label{fig:ISS}
\end{figure}

\textbf{Dynamic Resolution Variants.} To understand the upper limit of our architecture and pipeline, we also train VLMs with dynamic resolution schemes as a comparison. For such a purpose, we adopt a tiling strategy (Figure~\ref{fig:ISS}(b)) as in MiniCPM~\cite{hu2024minicpm} and utilize AIMv2~\cite{fini2024multimodal} for the vision encoder. This dynamic setting demonstrates a better performance of the current data and training pipeline and illustrates the gap over the default fixed version.

\textbf{Implicit Semantic Stitching.} 
In addition, we discover that the scheme of Figure~\ref{fig:ISS} (b) directly clips the semantically connected regions, breaking the semantic continuity among these image patches.
To mitigate this semantic discontinuity effect, a rather straightforward idea is to adopt overlapping alongside cropping as shown in Figure~\ref{fig:ISS} (c). Surprisingly, the inclusion of a large amount of repetitive information can degrade the performance, as detailed later in the ablation experiment in Section 4.8. 

\begin{figure}
    \centering
    \includegraphics[width=0.7\textwidth]{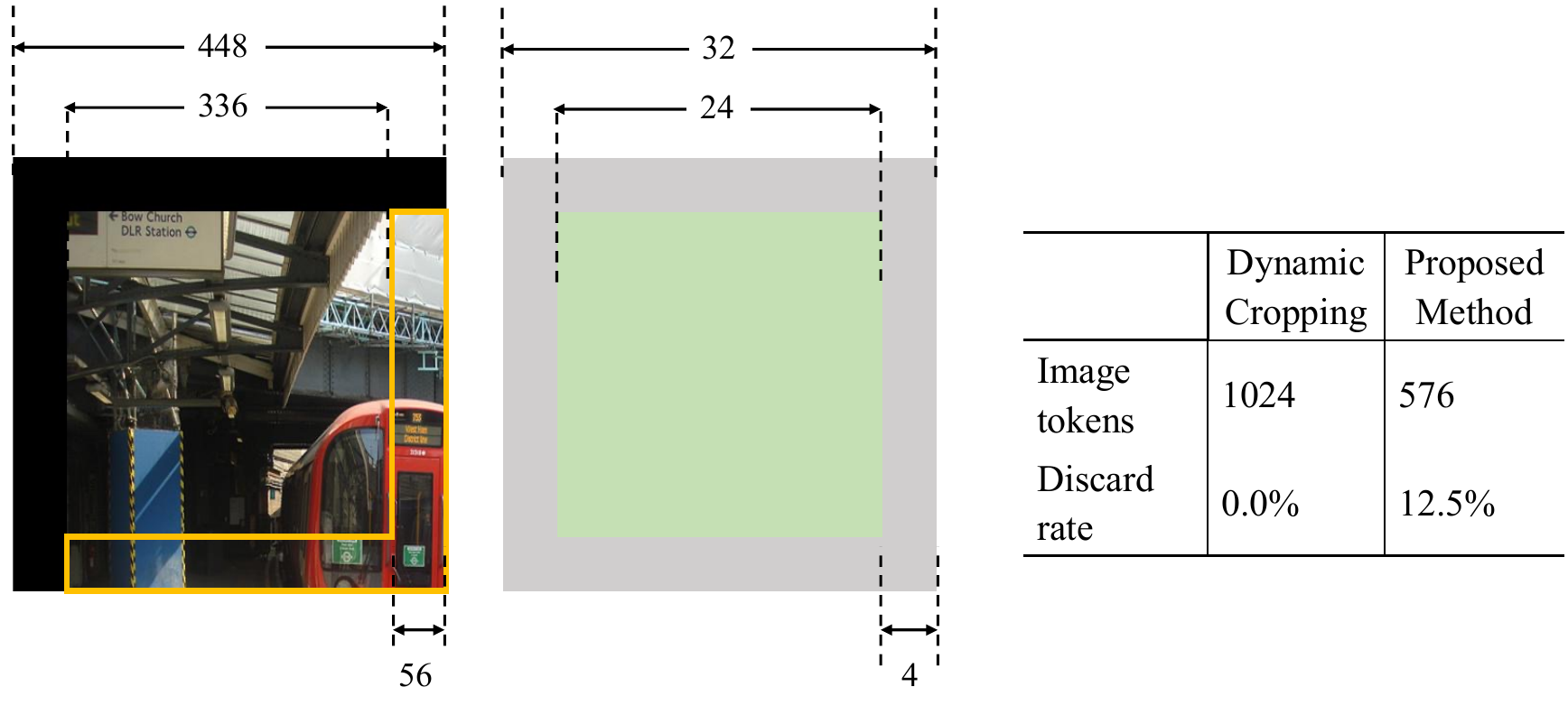}
    \caption{Implicit Semantic Stitching. \textbf{Left:} An example of one image tile and its feature map, where the yellow box represents the repetitive areas. The gray area represents the overlapping tokens, and the green ones represent retained tokens. \textbf{Right:} The overlap rate is approximately 12.5\%; The number of image tokens of each image tile is 576.}
    \label{fig:ISS_config}
    \vskip -0.2 in
\end{figure}

To address this issue, we propose \emph{Implicit Semantic Stitching} (ISS), which extracts image tiles' boundary consensus features as a type of inter-tile semantic glue, ensuring semantic consistency without sacrificing dynamic cropping benefits. As shown in Figure~\ref{fig:ISS} (d). We first pad the input image with a black frame to facilitate cropping. Then for image cropping, we reuse the overlapping scheme. After extracting the visual features, we remove the overlapping image tokens (the gray ones represent overlapping tokens, and the green ones represent retained tokens). Although all the visual tokens in the first image tile do not explicitly include boundary tokens from the second image tile, the boundary features of the second tile are implicitly incorporated via feature embedding. Metaphorically speaking, ISS stitches together the cutting-off semantic regions. The parameter details of the method are shown in Figure~\ref{fig:ISS_config}, the maximum number of image tile of one inputs image is limited to 4, for one image tile, the overlap rate is approximately 12.5\%, the number of image tokens is 576.

\subsection{Data Choices}
\begin{figure}[htbp]
  \begin{minipage}[t]{0.38\textwidth}
  \vspace{0pt}
    \includegraphics[width=\linewidth]{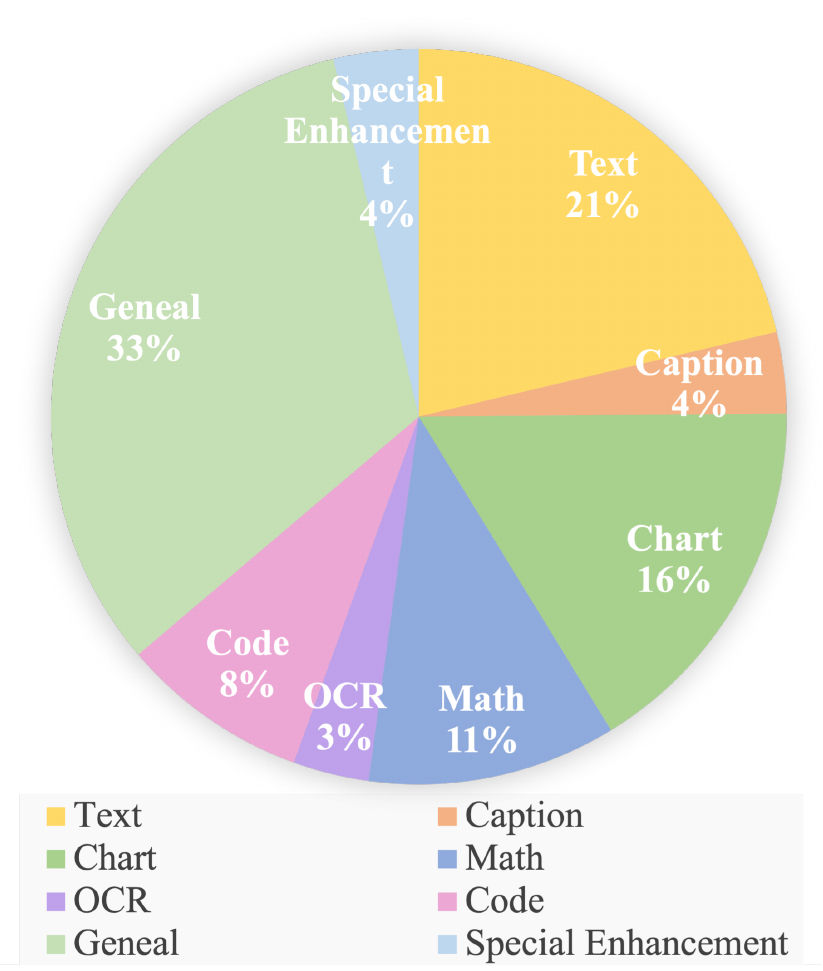}
    \caption{Pie chart of the Stage 4-10M data categorized by General, Code, OCR, Math, Chart, Caption, Text, and Special enhancement.}
    \label{fig:stage4-pie}
  \end{minipage}%
  \hspace{-1cm}%
  \begin{minipage}[t]{0.85\textwidth}
  \vspace{0pt}
    \centering
    \resizebox{0.6\columnwidth}{!}{
    \begin{tabular}{lc}
      \toprule
      Data Source  & Type \\
      \midrule
      ShareGPT4V\cite{chen2024sharegpt4v}  & Caption \\
      ChartQA\cite{masry2022chartqa}  & Chart \\
      DocVQA\cite{mathew2021docvqa}   & Chart \\
      InfographicVQA\cite{mathew2022infographicvqa}   & Chart \\
      Deepform\cite{svetlichnaya2020deepform}   & Chart \\
      Kleister\cite{stanislawek2021kleister}   & Chart \\
      TabFact\cite{chen2019tabfact} &   Chart \\
      WikiTableQuestions\cite{pasupat2015compositional}   & Chart \\
      VisualMRC\cite{tanaka2021visualmrc}   & Chart \\
      TextCaps\cite{sidorov2020textcaps}   & Chart \\
      TextVQA\cite{singh2019towards} &   Chart \\
      ScigraphQA\cite{li2023scigraphqa}   & Chart \\
      MultiMath\cite{peng2024multimath}   & Math \\
      MAVIS\cite{zhang2024mavis} &   Math \\
      DocStruct4M\cite{hu2024mplug}   & OCR \\
      Websight\cite{laurenccon2024unlocking}   & Code \\
      MMDU\cite{liu2024mmdu} &   Geneal \\
      M4Instruct\cite{li2024llavanext}   & Geneal \\
      M3IT\cite{li2023m3it} &   Geneal \\
      ALLaVA\cite{hardy2024allava}   & Geneal \\
      ShareGPT-4o\cite{laboratory2024sharegpt4o}   & Special Enhancement \\
      \bottomrule
    \end{tabular}
    }
    \caption{Multi-modal Data sources of Stage-4.}
    \label{tab:stage-4-data}
  \end{minipage}
\end{figure}

Data curation has been one of the core advances for VLM's performance. Among the large-scale open-source multimodal dataset collections available, LLaVA-One-Vision~\cite{li2024llava} and  InfinityMM~\cite{gu2024infinitymmsc} are the most prominent due to both quantity and quality. VLMs trained on them showed distinguished performance. To facilitate the reproducibility of our models, we build Flash-VL on InfinityMM (Stage 1 to 3) along with an independent Stage 4 constructed by assembling a large set of open-source datasets shown in Table~\ref{tab:stage-4-data} and Figure~\ref{fig:stage4-pie}. The preference dataset used in Stage 5, approximately 87k, is mainly curated from \cite{li2024vlfeedback} along with some partial in-house data.


\subsection{Training Strategies}

Our Flash-VL follows a multi-stage training pipeline, see Table~\ref{tab:multi-stage-comp}. During Stage 1, only the Adapter is set to be learnable to align vision representations with textual embeddings, and the learning rate decays from 1e-3 to 2e-5. For the Stage 2 to 4, all parameters are active for training and have a cosine annealing from 1e-5 to zero. All data samples are seen only once. For Stage 5, we adopt Direct Preference Optimization (DPO)~\cite{rafailov2023dpo} with LoRA~\cite{hu2022lora} only on the LLM and the Adapter.

\begin{table}[!ht]
\centering
\setlength{\tabcolsep}{3pt}
\begin{tabular}{lccccc}
\toprule
                         & Stage 1     & Stage 2    & Stage 3 & Stage 4 & Stage 5 \\
\midrule
Train Phase              & Pre-train   & Fine-tune & Fine-tune & Fine-tune & DPO \\
Resolution               & 512         & 512 & 512 & 512 & 512\\
Data Samples             & 10M         & 24M & 6M & 10M & 87k\\
Trainable                & Adapter   & Full Model & Full Model & Full Model & LLM+Adapter\\
Train Type               & Full        & Full       & Full       & Full  & LoRA \\
Trainable Params         & 91.02M      & 1733.07M & 1733.07M & 1733.07M  & 10.73M  \\
LR Schedule              & Cosine Decay&Cosine Decay &  Cosine Decay & Cosine Decay & Cosine Decay\\
LR                       & 1.00E-03    & 1.00E-05 & 1.00E-05 & 1.00E-05 & 1.00E-05 \\
$\text{LR}_{\text{min}}$ & 2.00E-05    & 0 & 0 & 0 & 0 \\
Epoch                    & 1           & 1 & 1 & 1 & 1 \\
\bottomrule
\end{tabular}
\vskip 0.1in
\caption{Configuration for training Flash-VL-2B across various stages.}
\label{tab:multi-stage-comp}
\vskip -0.2in
\end{table}

\section{Experiments}

\subsection{Setup}\label{sec:setup}
\textbf{Ablation Setting.} For ablation studies, we pretrain the models on a 2.5M combined samples of ShareGPT4V-PT-1246K~\cite{chen2024sharegpt4v}, ALLaVA-Caption-710K~\cite{chen2024allava}, LLaVA-CC3M-Pretrain-595K~\cite{liu2023llava}, and then finetune them on the LLaVA-1.5-665K~\cite{liu2023improvedllava} dataset. The pretraining takes 8$\times$ A100 GPUs to train, with a global batch size of 48, an initial learning rate of 1e-3, a minimum learning rate of 2e-5, and 400 steps of warm-ups. Finetuning starts with a learning rate of 1e-5 and decays to zero with cosine annealing. All data samples are seen only once.

\textbf{Benchmarks.} We evaluate our models and the state-of-the-art lightweight VLMs on a myriad of multimodal benchmarks such as MMMU~\cite{yue2024mmmu}, MMBench~\cite{MMBench}, MMStar~\cite{chen2024we}, MMVet~\cite{yu2023mm}, MathVista~\cite{lu2023mathvista}, AI2D~\cite{kembhavi2016diagram}, HallusionBench~\cite{guan2024hallusionbench}, OCRBench~\cite{liu2024ocrbench}, MME~\cite{fu2023mme}, SEEDBench~\cite{li2023seedbench}. For multimodal reasoning capacity, we evaluate on DynaMath~\cite{zou2024dynamath}, MathVision~\cite{wang2024measuring}, MathVerse~\cite{zhang2024mathverse}, MMMU Pro~\cite{yue2024mmmupro}, WeMath~\cite{qiao2024wemathdoeslargemultimodal}. We adopt  VLMEvalkit~\cite{duan2024vlmevalkit} for the evaluation purpose of all models.


\textbf{Full Pipeline Setup.} Table~\ref{tab:multi-stage-comp} gives the hyperparameters for our complete pipeline training. For stage 5, we grid search the $\beta$ of DPO to find 0.1 gives the best performance. It comprises five stages. We train our models on 40 $\times$ A100-80G GPUs with a global batch size of 240. Note we choose SigLIP2-so400m for the static resolution version and AIMv2 H/14\@448 for the dynamic variants as we find the dynamic version of AIMv2 shows better performance than SigLIP2 under the ablation setting, see Table~\ref{tab:dynamic-comp} of Section~\ref{app:ablation-dyn}.

\subsection{Comparison of state-of-the-art lightweight MLLMs}

Table~\ref{tab:vlm-comp} gives the comparison with Qwen2-VL, InternVL2.5, and Aquila-VL (all in the range of 2B parameters) on 11 common VLM benchmarks. These benchmarks span a wide range of categories, seeking to ensure a rigorous and balanced evaluation of Flash-VL’s performance in various multimodal tasks.
First, as shown in Table~\ref{tab:vlm-comp}, Flash-VL-2B$_{\text{d-ISS}}$ surpasses previous state-of-the-art method InternVL2.5-2B by 1.2\% on average. Specifically, on MathVista, Flash-VL-2B$_{\text{d-ISS}}$ outperforms the previous state-of-the-art method Aquila-VL-2B by 2.1\%. On OCRBench, Flash-VL-2B$_{\text{d-ISS}}$ outperforms the previous state-of-the-art method InternVL2.5-2B by 43. These results showcase the ability of Flash-VL-2B$_{\text{d-ISS}}$ to handle general multimodal understanding and reasoning tasks.
Second, Compared with Flash-VL-2B$_{\text{d}}$, Flash-VL-2B$_{\text{d-ISS}}$ improves it by 3.4\%, 3.5\%, 12 and 0.8\% for MathVista$_{\text{testmini}}$, HallusionBench, OCRBench, and average of 11 benchmarks respectively, which demonstrates the advantage of the ISS technique in enhancing general multimodal tasks.
Third, Flash-VL-2B$_{\text{s}}$ with higher throughput outperforms Qwen2-VL-2B on average of 11 benchmarks by 2.2\%. For instance, Flash-VL-2B$_{\text{s}}$ delivers significant gains on MMBench$^{\text{en}}$  (78.4 vs. 74.9), MMStar (53.8 vs. 48.0), and MathVista$_{\text{testmini}}$ (59.3 vs. 43.0), these results further indicate the great potential of Flash-VL-2B$_{\text{s}}$.

\begin{table}[!ht]
\begin{center}
\setlength{\tabcolsep}{2pt}
\resizebox{1.01\columnwidth}{!}{
\begin{tabular}{lH*{6}{|c}}
\toprule
Benchmark & InternVL2-2B  & Qwen2-VL-2B & Aquila-VL-2B & InternVL2.5-2B & Flash-VL-2B$_{\text{s}}$ & Flash-VL-2B$_{\text{d}}$ & Flash-VL-2B$_{\text{d-ISS}}$\\
\midrule
MMMU$_{\text{val}}$            & 36.3  & 41.9  & 44.4  & 41.8 & 43.6   & 42.9 & 42.9 \\
MMBench$^{\text{en}}$          & 71.2  & 74.9  & 78.6  & 74.7 & 78.4   & 78.4 & 79.1 \\
MMBench$^{\text{cn}}$          & 70.9  & 73.5  & 76.3  & 71.6 & 74.7   & 74.9 & 76.7 \\
MMStar                         & 49.8  & 48.0  & 54.9  & 54.1 & 53.8   & 54.4 & 54.1 \\
MathVista$_{\text{testmini}}$  & 37.7  & 43.0  & 59.4  & 50.9 & 59.3   & 58.1 & 61.5 \\
AI2D$_{\text{test}}$           & 64.1  & 74.1  & 75.0  & 75.1 & 74.2   & 74.1 & 74.4 \\
MMVet                          & 39.5  & 49.5  & 40.9  & 61.7 & 47.3   & 52.7 & 50.7 \\
HallusionBench                 & 36.8  & 39.2  & 38.5  & 42.7 & 43.5   & 45.5 & 49.0 \\
OCRBench                       & 784   &  794  &  773  & 800  & 764    & 831  & 843  \\
MME                            & 1876  & 1872  & 1813  & 2091 & 1715   & 1866 & 1850 \\
SEEDBench                      & 71.6  & 71.5  & 78.9  & 73.2 & 73.6   & 73.6 & 74.5 \\
\midrule                 
Average                        & 56.7  & 60.2  & 62.6  & 63.6 & 62.4   & 64.0 & 64.8\\
\bottomrule
\end{tabular}
}
\end{center}
\caption{Performance comparison with state-of-the-art lightweight MLLMs on 11 common VLM benchmarks. For average score calculation, MME's score is normalized by dividing it with 28 and OCRBench's score is divided by 10 (same for below). ${\text{s}}$: static resolution, ${\text{d}}$: dynamic tiling w/o ISS, ${\text{d-ISS}}$: dynamic tiling w/ ISS.}
\label{tab:vlm-comp}
\vskip -0.2in
\end{table}

\subsection{Comparison of ViTs}

We give the VLM performance comparison under the ablation setting (Section ~\ref{sec:setup}) in Table~\ref{tab:vit-comp} when choosing AIMv2~\cite{fini2024multimodal} and SigLIP2~\cite{tschannen2025siglip} as visual encoders, all with static input resolutions. SigLIP2 generally demonstrates a richer power of visual representation, which transfers well on multiple VLM tasks. As shown in Table~\ref{tab:vit-comp}, SigLIP2-so400m@512 achieves the state-of-the-art performance with an average score of 50.0\%, a +3.10\% improvement over AIMv2-H/14@448, surpassing SigLIP2-L/16@384. Notably, on OCRBench, SigLIP2-so400m@512 achieves leading performance compared with AIMv2-H/14@448 (e.g. 386 vs. 184). We also conduct ablation experiments on dynamic resolution when choosing AIMv2 and SigLIP2 as visual encoders in Table~\ref{tab:dynamic-comp} of Section~\ref{app:ablation-dyn}, we find that dynamic resolution has a negative impact when using SigLIP2, which is left for future study.

\begin{table}[!ht]
\begin{center}
\begin{tabular}{l|c|c|c}
\toprule
Benchmark & AIMv2-H/14@448 & SigLIP2-L/16@384 & SigLIP2-so400m@512 \\
\midrule
visual tokens  & 256 & 144 & 256 \\
\midrule
MMMU$_{\text{val}}$                & 40.3   & 38.8  &  38.8 ~~(-1.5)\\
MMBench$^{\text{en}}$              & 68.2   & 69.0  &  71.7 ~~(+3.5)\\
MMBench$^{\text{cn}}$              & 66.4   & 67.5  &  68.2 ~~(+1.8)\\
MMStar                             & 43.3   & 41.5  &  43.3 ~~(+0.0)\\
MathVista$_{\text{testmini}}$      & 27.9   & 27.8  &  27.9 ~~(+0.0)\\
AI2D$_{\text{test}}$               & 60.0   & 60.1  &  62.3 ~~(+2.3)\\
MMVet                              & 33.8   & 32.4  &  34.4 ~~(+0.6)\\
HallusionBench                     & 34.8   & 38.3  &  39.8 ~~(+5.0)\\
OCRBench                           & 184    & 342   &  386  ~~(+202)~\\
MME                                & 1505   & 1557  &  1531 ~(+26)\\
SEEDBench                          & 69.1   & 69.1  &  70.7 ~~(+1.6)\\
\midrule
Average                            & 46.9   & 48.6   & 50.0 ~~(+3.1)\\
\bottomrule
\end{tabular}
\end{center}
\caption{Comparison with various vision encoders on 11 common VLM benchmarks. +/- are based on AIMv2-H/14@448.}
\label{tab:vit-comp}
\end{table}

\subsection{Ablation Study of DPO}
As shown in Table \ref{tab:dpo-comp}, Flash-VL-2B$_{\text{s}}$, Flash-VL-2B$_{\text{d}}$, and Flash-VL-2B$_{\text{d-ISS}}$ all achieve superior performance compared with their pre-DPO counterparts. Notably, on OCRBench, the score rises from 696 to 764 for Flash-VL-2B$_{\text{s}}$, from 786 to 831 for Flash-VL-2B$_{\text{d}}$, and from 795 to 843 on the Flash-VL-2B$_{\text{d-ISS}}$. As per MMVet, the score significantly improves from 40.4 to 47.3 for Flash-VL-2B$_{\text{s}}$, from 48.1 to 52.7 for Flash-VL-2B$_{\text{d}}$, and from 45.5 to 50.7 for Flash-VL-2B$_{\text{d-ISS}}$.
These results highlight the effectiveness of the curated DPO dataset and the training scheme in enhancing multimodal capabilities. 

\newcolumntype{G}{>{\columncolor[gray]{0.9}\centering\arraybackslash}m{1.8cm}}
\newcolumntype{H}{>{\columncolor[gray]{0.9}\centering\arraybackslash}m{1.9cm}}

\begin{table}[!ht]
\begin{center}
\begin{small}
\setlength{\tabcolsep}{1pt}
\resizebox{1.01\columnwidth}{!}{
\begin{tabular}{l|c|G|c|G|c|H}
\toprule
Benchmark 
& Flash-VL-2B$_{\text{s}}$ 
& Flash-VL-2B$_{\text{s}}$-DPO  
& Flash-VL-2B$_{\text{d}}$ 
& Flash-VL-2B$_{\text{d}}$-DPO 
& Flash-VL-2B$_{\text{d-ISS}}$ 
& Flash-VL-2B$_{\text{d-ISS}}$-DPO \\
\midrule
MMMU$_{\text{val}}$                & 40.8  & 43.6 ~(+2.8)       & 43.3   & 42.9 ~~(-0.4)     & 40.7     & 42.9 ~~(+2.2) \\
MMBench$^{\text{en}}$              & 76.7  & 78.4 ~(+1.7)       & 76.5   & 78.4 ~~(+1.9)     & 77.8     & 79.1 ~~(+1.3) \\
MMBench$^{\text{cn}}$              & 74.1  & 74.7 ~(+0.6)       & 74.7   & 74.9 ~~(+0.2)     & 75.0     & 76.7 ~~(+1.7) \\
MMStar                             & 53.4  & 53.8 ~(+0.4)       & 54.3   & 54.4 ~~(+0.1)     & 55.0     & 54.1 ~~(-0.9) \\
MathVista$_{\text{testmini}}$      & 56.1  & 59.3 ~(+3.2)       & 54.7   & 58.1 ~~(+3.4)     & 62.0     & 61.5 ~~(-0.5) \\
AI2D$_{\text{test}}$               & 73.6  & 74.1 ~(+0.5)       & 74.1   & 74.1 ~~(+0.0)     & 75.2     & 74.4 ~~(-0.8) \\
MMVet                              & 40.4  & 47.3 ~(+6.9)       & 48.1   & 52.7 ~~(+4.6)     & 45.5     & 50.7 ~~(+5.2) \\
HallusionBench                     & 44.6  & 43.5 ~(-1.1)       & 43.4   & 45.5 ~~(+2.1)     & 47.6     & 49.0 ~~(+1.4) \\
OCRBench                           & 696   & 764  ~~(+68)       & 786    & 831  ~~(+45)      & 795      & 843 ~~(+48) \\
MME                                & 1804  & 1715 ~~(-89)        & 1910   & 1866 ~~(-44)      & 1894     & 1850 ~~(-44) \\
SEEDBench                          & 73.2  & 73.6 ~(+0.4)       & 73.3   & 73.6 ~~(+0.3)     & 74.4     & 74.5 ~~(+0.1) \\
\midrule           
Average                            & 60.6  & 62.4 ~(+1.8)       & 62.7   & 64.0 ~(+1.3)      & 63.7     & 64.8 ~~(+1.1) \\
\bottomrule
\end{tabular}
}
\end{small}
\end{center}
\caption{Ablation study using DPO dataset on Flash-VL-2B$_{\text{s}}$, Flash-VL-2B$_{\text{d}}$ and Flash-VL-2B$_{\text{d-ISS}}$. }
\label{tab:dpo-comp}
\vskip -0.2in
\end{table}

\subsection{Multi-stage Training}

Table~\ref{tab:multi-stage} illustrates the performance after each stage of training Flash-VL-2B with the static resolution strategy. We observe that the average scores indicate a consistent improvement in performance as the training data scales. Note that as we gradually introduce higher-quality data and more challenging questions, the overall performance can be consequently enhanced.

\begin{table}[!ht]
\begin{center}
\begin{tabular}{l|c|c|c|c|c}
\toprule
Benchmark                     & Stage 1& Stage 2& Stage 3 & Stage 4 & Stage 5\\
\midrule
MMMU$_{\text{val}}$           & 38.8   & 41.3  & 42.0  & 40.8  & 43.6  \\
MMBench$^{\text{en}}$         & 67.4   & 70.2  & 74.4  & 76.7  & 78.4  \\
MMBench$^{\text{cn}}$         & 66.1   & 65.8  & 70.9  & 74.1  & 74.7   \\
MMStar                        & 41.9   & 44.6  & 51.5  & 53.4  & 53.8  \\
MathVista$_{\text{testmini}}$ & 27.8   & 45.7  & 55.1  & 56.1  & 59.3   \\
AI2D$_{\text{test}}$          & 62.2   & 66.9  & 72.9  & 73.6  & 74.1  \\
MMVet                         & 32.6   & 38.3  & 38.0  & 40.4  & 47.3  \\
HallusionBench                & 38.4   & 39.7  & 39.6  & 44.6  & 43.5  \\
OCRBench                      & 365    & 563   & 685   & 696   & 764    \\
MME                           & 1499   & 1623  & 1778  & 1804  & 1715  \\
SEEDBench                     & 69.9   & 71.3  & 72.5  & 73.2  & 73.6  \\
\midrule
Average                       & 48.6   & 54.4 & 59.0   & 60.6  & 62.4 \\
\bottomrule
\end{tabular}
\end{center}
\caption{Data Scaling Performance of Flash-VL-2B at multiple training stages. The checkpoint at Stage 1 is fine-tuned with LLaVA-1.5-665K to have instruction following capabilities.}
\label{tab:multi-stage}
\vskip -0.2in
\end{table}

\subsection{Comparison of Different Image Cropping Strategies}

Table \ref{tab:crop-comp} compares the VLM performance of different image cropping methods mentioned in Figure~\ref{fig:ISS} under the ablation setting (Section~\ref{sec:setup}). All experiments adopt AIMv2 (H/14@448) as the vision encoder and Qwen-2.5-1.5B as the language model. We can find that the overlapping cropping strategy drops by 0.80\% compared with the non-overlapping version. In contrast, our proposed ISS can further improve by 0.7\% compared with dynamic cropping. For the full training of Flash-VL, the same strategy helps Flash-VL-2B$_{\text{d-ISS}}$ to improve by 0.8\% as shown in Table~\ref{tab:vlm-comp}. Notably, the dynamic overlapping cropping strategy severely impaired the OCR and document understanding capabilities (383 vs. 366 on OCRBench). In contrast, ISS increases OCRBench from 383 to 427. This observation suggests that a large amount of repetitive information can degrade the performance, while ISS can mitigate the shortcomings of dynamic cropping by providing semantic consistency.

\begin{table}[ht]
\begin{center}
\begin{tabular}{m{3cm}|m{2cm}|m{2cm}m{2cm}m{2cm}}
\toprule
Method & Static  & Dynamic Cropping & Dynamic Overlapping Cropping  &  Implicit Semantic Stitching \\
\midrule
MMMU$_{val}$           & 40.3   & 39.6  & 39.0 ~~(-0.6)  & 40.4 ~~(+0.8)  \\
MMBench$^{en}$         & 68.2   & 71.8  & 70.1 ~~(-0.7)  & 72.8 ~~(+1.0)  \\
MMBench$^{cn}$         & 66.4   & 70.4  & 68.4 ~~(-2.0)  & 70.2 ~~(-0.2)  \\
MMStar                 & 43.3   & 42.2  & 44.9 ~~(+2.7)  & 45.1 ~~(+2.9)   \\
MathVista$_{testmini}$ & 27.9   & 28.4  & 27.6 ~~(-0.8)  & 28.5 ~~(+0.1)  \\
AI2D$_{test}$          & 60.0   & 63.2  & 62.1 ~~(-1.1)  & 63.2 ~~(+0.0)  \\
MMVet                  & 33.8   & 35.2  & 33.8 ~~(-1.4)  & 33.9 ~~(-1.3)  \\
HallusionBench         & 34.8   & 36.3  & 36.7 ~~(+0.4)  & 35.3 ~~(-1.0)    \\
OCRBench               & 184    & 383   & 366  ~~~(-17)  & 427  ~~~(+44)    \\
MME                    & 1505   & 1697  & 1601 ~~(-96)   & 1687 ~~(-10)     \\
SeedBench              & 69.1   & 71.4  & 72.1 ~~(+0.7)  & 72.5 ~~(+1.1)  \\
\midrule
Average                & 46.9   & 50.6  & 49.9 ~(-0.7)          & 51.4 ~(+0.8)  \\
\bottomrule
\end{tabular}
\end{center}
\caption{Comparison of various image processing methods. +/- are based on Dynamic Dropping.}
\label{tab:crop-comp}
\vskip -0.2in
\end{table}

\subsection{Comparison of Multimodal Reasoning Capacity}~\label{sec:comp-multimodal-reason}
We conduct a series of comparative experiments to explore the effect of lightweight supervised fine-tuning (SFT) and reinforcement learning (RL) on enhancing the model's reasoning capabilities without relying on a long chain of thoughts. We use the stage 4 model of Flash-VL-2B$_{\text{s}}$ as the baseline and evaluate the effect of SFT,  RL, and their combination on top of it. Specifically, we use LoRA~\cite{hu2022lora} for SFT and GRPO~\cite{shao2024deepseekmath} for RL. We evaluate the trained models on multiple mainstream mathematical and scientific reasoning benchmark datasets, with the average performance and subcategory results summarized in Table \ref{tab:math_benchmark_results_generated}. More training settings are detailed in Section~\ref{app:reasoning_setup}. 

We find that both lightweight supervised fine-tuning and reinforcement learning, when applied individually, significantly improve the model's performance on mathematical reasoning tasks. Moreover, combining SFT and RL further enhances the model's reasoning ability. The combination of SFT and RL lead to optimal average performance across all major mathematical reasoning benchmarks, effectively demonstrating the complementarity and synergistic benefits of these two methods. 

\begin{table}[htbp]
\centering
\setlength{\tabcolsep}{3pt}
\begin{tabular}{l|c|ccccc}
\toprule
Method & Average & DynaMath & MathVision & MathVerse & MMMU Pro & WeMath \\
\midrule
Flash-VL-2B$_{\text{s}}$  & 23.80       & 23.19       & 26.72       & 16.84       & 16.24        & 36.03   \\
+ SFT                    & 26.08 (+2.28) & 28.28       & 31.06       & 16.97       & 15.95        & 38.16   \\
+ RL                     & 27.23 (+3.43) & 26.94       & 27.94       & 17.73       & \textbf{16.99}        & 46.55   \\
+ SFT + RL               & \textbf{29.05 (+5.25)} & \textbf{30.61} & \textbf{32.48} &  \textbf{18.45}    & 16.53         & \textbf{47.18}            \\
\bottomrule
\end{tabular}
\vskip 0.1in
\caption{Comparison of Flash-VL-2B$_{\text{s}}$ and its variants on mathematical and  reasoning benchmarks.}
\label{tab:math_benchmark_results_generated}
\vskip -0.1in
\end{table}

\subsection{Comparison on Latency and Throughput}
We compare our model with state-of-the-art models in terms of latency and throughput. We benchmark all the models on an NVIDIA L40 GPU using \texttt{transformers} with all samples from MMMU and the number of output tokens set to 128. The result is shown in Figure~\ref{fig:pareto}. Thanks to the architecture choices and image encoding strategies, Flash-VL-2B$_{\text{s}}$ leads in the throughput, striking the highest tokens per second (60.73). The whole Flash-VL-2B series forms a new Pareto-front in the realm of VLM models of 2B parameters. We also give the comparison of latency distributions in \ref{app:ttft_tpot}.

\section{Conclusion}
Motivated by stringent deployment requirements for VLMs, we construct a lightweight Flash-VL to strike a trade-off between latency and performance. Specifically, we exploit compact visual representation, effective token compression, and a powerful LLM component to build up our VLM. We also make use of the high-quality open-source multi-modal data collection as our training data. We adopt a multi-stage training pipeline that enhances its abilities gradually. Our competitive results attests that by adopting open-source models and datasets with latency constraints in mind, we can produce ultra-fast models with high performance. We hope our release could help advance the field of lightweight MLLMs.

\newpage

{
\small
\bibliographystyle{plain}
\bibliography{arxiv}
}

\newpage

\appendix

\section{Additional Experiments and Ablations}

\subsection{Comparison of Adapters}\label{app:ablation-adp}
Current VLMs simply utilize two MLPs (i.e. linear layer) for the adapter. We compare our proposed adapter with this simplistic design, shown in Table~\ref{tab:adp-comp}. Both adapters use pixel shuffling to compress visual tokens first. The same in-house ViT-Huge and Qwen-2-1.5B are used for ViT and LLM, respectively.

\begin{table}[!ht]
\begin{center}
\begin{tabular}{l|c|c}
\toprule
Benchmark & 2 $\times$ Linear & Proposed Adapter \\
\midrule
MMMU$_{\text{val}}$   & 36.7   & 38.3 \\
MMBench$^{\text{en}}$ & 56.5   & 61.5 \\
MMBench$^{\text{cn}}$ & 55.2   & 57.3\\
MMStar         & 33.6   & 35.7 \\
MathVista$_{\text{testmini}}$      & 27.8   & 24.0 \\
AI2D$_{\text{test}}$  & 55.2   & 54.6 \\
MMVet          & 17.6   & 23.9 \\
HallusionBench & 27.6   & 33.7 \\
OCRBench       & 54     & 168 \\
MME            & 1436   & 1569 \\
SEEDBench      & 55.1   & 62.6\\
\midrule
Average        & 39.4   & \textbf{40.8} \\
\bottomrule
\end{tabular}
\end{center}
\caption{VLM performance comparison of two adapter designs.}
\label{tab:adp-comp}
\end{table}

\subsection{Comparison of LLMs}\label{app:ablation-llm}

With an in-house contrastively-trained ViT-Huge as the same vision encoder, we evaluate the performance of various LLMs, namely LLaMA-3.2-1B (1.23B parameters), Qwen-2-1.5B and Qwen-2.5-1.5B, under the same framework and the ablation data setting. Among three candidate lightweight LLMs, Qwen-2.5-1.5B demonstrates superiority by a clear margin, shown in Table~\ref{tab:llm-comp}. Hence for the rest of experiments throughout the paper, we adopt Qwen-2.5-1.5B as the default LLM component. 

\begin{table}[!ht]
\begin{center}
\begin{tabular}{l|c|c|c}
\toprule
Benchmark & LLaMA-3.2-1B & Qwen-2-1.5B & Qwen-2.5-1.5B \\
\midrule
MMMU$_{\text{val}}$   & 30.3 & 34.8  & 38.3 \\
MMBench$^{\text{en}}$ & 41.4 & 62.3  & 61.5 \\
MMBench$^{\text{cn}}$ & 42.4 & 56.7  & 57.3\\
MMStar         & 28.5 & 35.0  & 35.7 \\
MathVista$_{\text{testmini}}$      & 23.5 & 15.9  & 24.0 \\
AI2D$_{\text{test}}$  & 54.0 & 52.3  & 54.6 \\
MMVet          & 20.1 & 18.2  & 23.9 \\
HallusionBench & 29.1 & 28.4  & 33.7 \\
OCRBench       & 193  &  195  & 168 \\
MME            & 1324 & 1477  & 1569 \\
SEEDBench      & 46.9 & 62.1  & 62.6\\
\midrule
Average        & 33.2 & 38.3  & \textbf{40.8} \\
\bottomrule
\end{tabular}
\end{center}
\caption{VLM performance comparison under various lightweight LLM choices.}
\label{tab:llm-comp}
\vskip -0.2in
\end{table}

\subsection{Ablation of Dynamic Resolution Setting on AIMv2 and SigLIP2}~\label{app:ablation-dyn}
Dynamic resolution, as a commonly used technique, can consistently improve model performance. We conducted ablation experiments on this technique when choosing AIMv2 and SigLIP2 as visual encoders, as shown in Table~\ref{tab:dynamic-comp}. Surprisingly, we discover that the dynamic setting hurts the performance on SigLIP2 while the AIMv2's version gets substantially enhanced. We leave it here as an open discussion why feeding more visual tokens is not more effective for SigLIP2.

\begin{table}[!ht]
\begin{center}
\resizebox{1.01\columnwidth}{!}{
\begin{tabular}{l|cc|cc}
\toprule
Benchmark & AIMv2 H/14@448 & + Dynamic Resolution & SigLIP2 so400m@512 & + Dynamic Resolution\\
\midrule
Max Image Split  & -1 & 4 & -1 & 4 \\
\midrule
MMMU$_{\text{val}}$                & 40.3   & 39.6  & 38.8  & 39.1         \\
MMBench$^{\text{en}}$              & 68.2   & 71.8  & 71.7  & 70.5        \\
MMBench$^{\text{cn}}$              & 66.4   & 70.4  & 68.2  & 68.4        \\
MMStar                             & 43.3   & 42.2  & 43.3  & 41.8        \\
MathVista$_{\text{testmini}}$      & 27.9   & 28.4  & 27.9  & 27.1        \\
AI2D$_{\text{test}}$               & 60.0   & 63.2  & 62.3  & 61.0        \\
MMVet                              & 33.8   & 35.2  & 34.4  & 32.3        \\
HallusionBench                     & 34.8   & 36.3  & 39.8  & 35.4        \\
OCRBench                           & 184    & 383   & 386   & 367         \\
MME                                & 1505   & 1697  & 1531  & 1662        \\
SEEDBench                          & 69.1   & 71.4  & 70.7  & 70.8         \\
\midrule
Average                            & 46.9   & 50.7  & 50.0  & 49.3           \\
\bottomrule
\end{tabular}
}
\end{center}
\caption{Abation of dynamic resolution with various vision encoders on 11 common VLM benchmarks. }
\label{tab:dynamic-comp}
\end{table}

\subsection{Comparison of TTFT/TPOT on MMMU}~\label{app:ttft_tpot}
\begin{figure}[htbp]
    \centering
    \begin{minipage}[t]{0.48\textwidth}
    \centering
    \includegraphics[width=7cm]{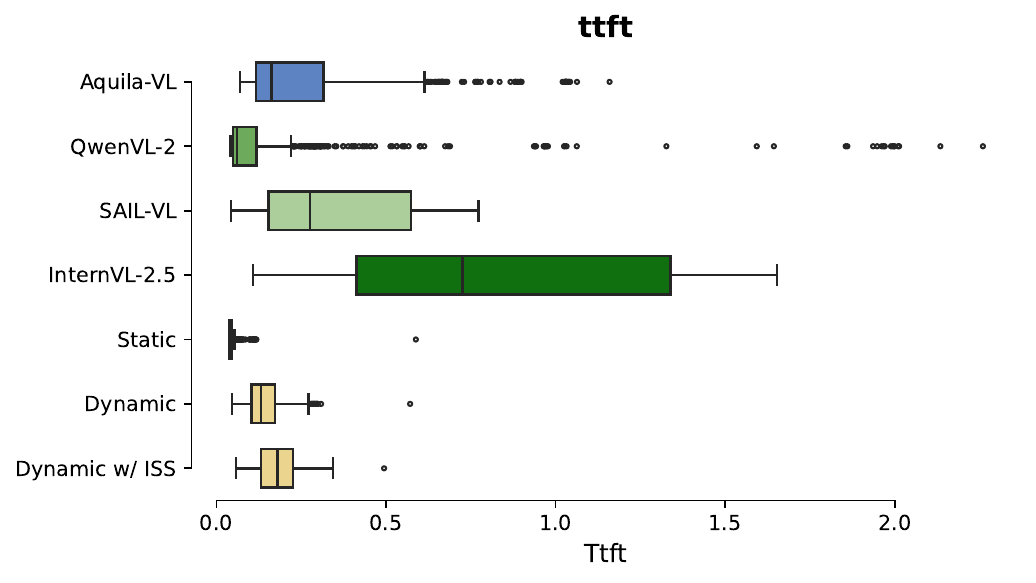}
    \end{minipage}
    \begin{minipage}[t]{0.48\textwidth}
    \centering
    \includegraphics[width=7cm]{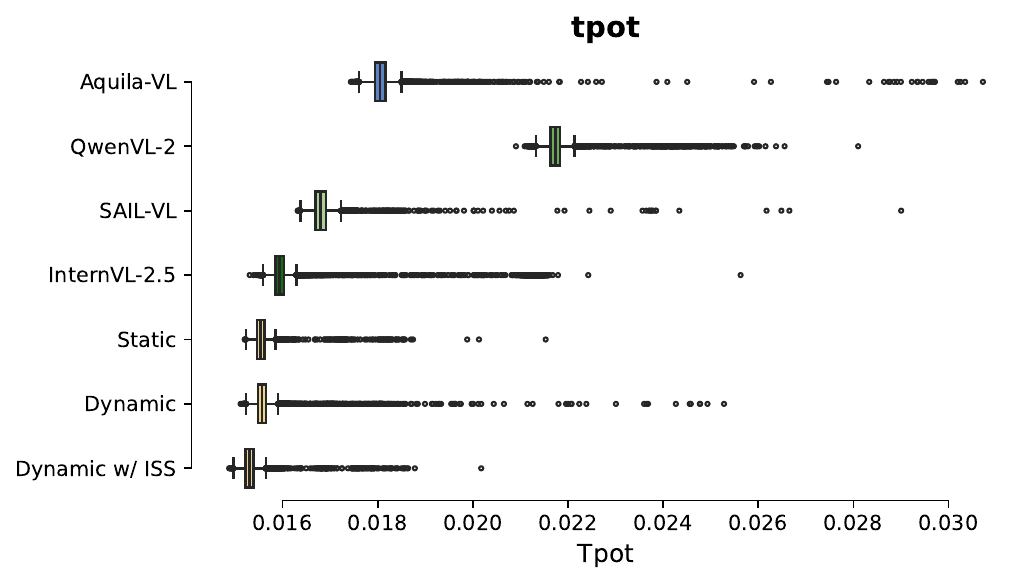}
    \end{minipage}
    \caption{\textbf{Left:} Distribution of TTFT on MMMU. \textbf{Right:} Distribution of TPOT on MMMU.}
    \label{fig:ttft-tpot}
\end{figure}

Apart from benchmarking the throughputs in Figure~\ref{fig:pareto}, we also profile the time-to-first-token (TTFT) and the time-per-output-token (TPOT) of the state-of-the-art 2B VLM models on MMMU~\cite{yue2024mmmu} with the official evaluation prompts in Figure~\ref{fig:ttft-tpot}. The number of output tokens is all set to 128. The former metric describes the prefilling latency, while the latter is about self-decoding speed. Our models exhibit the best TPOT on average. Although QwenVL-2-2B's image processing strategy benefits a competitive TTFT (0.14s), its wider range of embedding tokens contributes to a longer decoding time, effectively slowing down the overall throughput (39.07 tokens/s) as shown in Figure~\ref{fig:pareto}. Thanks to our token mechanisms and the choice of a compact vision encoder, our Flash-VL-2B$_{s}$, Flash-VL-2B$_{d}$, and Flash-VL-2B$_{d_{ISS}}$ enjoy the overall best average latencies (0.04, 0.13, 0.18 seconds on TTFT and approximately 0.015 seconds for all models on TPOT) and throughputs (60.73, 51.53, 48.66 tokens/s) out of all state-of-the-art models.

\section{Experimental Setup for Reasoning Capacity Comparison}
\label{app:reasoning_setup}

Here we detail the datasets and training strategies 
 in Section~\ref{sec:comp-multimodal-reason}.

For the SFT phase, we use the open-source distilled dataset Chinese-DeepSeek-R1-Distill-data-110k~\cite{Chinese-Data-Distill-From-R1}. This dataset primarily targets domains such as logical and mathematical reasoning, while also encompassing a proportion of general-purpose corpora. During training, only the \texttt{content} field was considered as the training label, while the \texttt{reasoning\_content} field was not utilized. We conducted SFT with the Low-Rank Adaptation (LoRA) ~\cite{hu2022lora} for efficient fine-tuning. The learning rate was set to $1 \times 10^{-5}$, coupled with a cosine annealing learning rate scheduler.

For the RL phase, we employ the open-source multimodal-open-r1-8k-verified~\cite{lmms-lab-multimodal-open-r1-8k} dataset. This is a visual multimodal dataset oriented towards geometric reasoning. The RL phase adopted  GRPO~\cite{shao2024deepseekmath}. Each group consists of 24 samples. The regularization coefficient $\beta$ for KL divergence is set to 0.001, and the learning rate is $5 \times 10^{-7}$ with a cosine annealing scheduler.

Both phases were trained for a single epoch each. The last checkpoint from the complete training was selected for final testing.

\section{Limitations}~\label{app:limit}
Although our models achieve state-of-the-art performance, which well balances the performance and speed, our overall performance might be limited by not exploiting a much larger set of in-house datasets as many closed-sourced VLMs do. Besides, the choice of standard benchmarks may not be comprehensive enough to reflect the full spectrum of VLM's capacity. In the meantime, we leave multi-image, video processing capabilities for the next generation and don't compare such scenarios. We also didn't target specific hardware devices, which may still introduce difficulties when it comes to real applications. Further, industry-level inference frameworks like vLLM~\cite{kwon2023efficient} or SGLang~\cite{sgl_project_sglang} could also be exploited to enhance the inference speed.

\end{document}